\documentclass[conference]{IEEEtran}
\IEEEoverridecommandlockouts
\usepackage{cite}
\usepackage{amsmath,amssymb,amsfonts}
\usepackage{algorithmic}
\usepackage{times}  
\usepackage{helvet}  
\usepackage{courier}  
\usepackage[hyphens]{url}
\usepackage{graphicx}
\usepackage{textcomp}
\usepackage{multirow}
\usepackage{adjustbox}
\usepackage{cuted} 
\usepackage{float}
\usepackage{caption}
\usepackage{amsmath}
\usepackage{array}
\usepackage{xcolor}
\begin{document}

\title{PolarGuide-GSDR: 3D Gaussian Splatting Driven by Polarization Priors and Deferred Reflection for Real-World Reflective Scenes}

\author{Derui Shan\textsuperscript{†,1},  Qian Qiao \textsuperscript{†,2}, Hao Lu \textsuperscript{3}, Tao Du \textsuperscript{*, 1},  Peng Lu \textsuperscript{*,2} \\
\textsuperscript{†}Equal contribution, \textsuperscript{*}Corresponding authors.\\
\textsuperscript{1}North China University of Technology, Beijing, China\\
\textsuperscript{2}Beijing University of Posts and Telecommunications, Beijing, China\\
\textsuperscript{3}University of Science and Technology Beijing, Beijing, China}

\maketitle
\begin{strip}
	\centering
	\includegraphics[width=\linewidth]{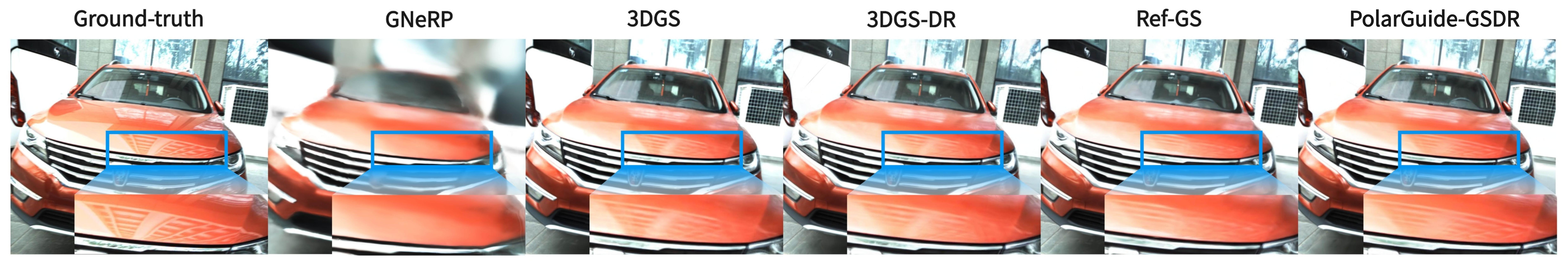}
	\vspace{-2em} 
	\captionof{figure}{Novel view synthesis results with specular reflections. From left to right: ground-truth, GNeRF~\cite{yang2024gnerp}, 3DGS~\cite{kerbl2023gaussian}, 3DGS-DR~\cite{ye2024deferredreflection}, Ref-GS~\cite{zhang2025refgs}, and ours. By introducing polarization information for guidance and supervision, our method achieves higher-quality reflection reconstruction.}
	\label{fig:first}
	\vspace{-1.7em} 
\end{strip}
\begin{abstract}
Polarization-aware Neural Radiance Fields (NeRF) enable novel view synthesis of specular-reflection scenes but face challenges in slow training, inefficient rendering, and strong dependencies on material/viewpoint assumptions. However, 3D Gaussian Splatting (3DGS) enables real-time rendering yet struggles with accurate reflection reconstruction from reflection-geometry entanglement, adding a deferred reflection module introduces environment map dependence. We address these limitations by proposing PolarGuide-GSDR, a polarization-forward-guided paradigm establishing a bidirectional coupling mechanism between polarization and 3DGS: first 3DGS’s geometric priors are leveraged to resolve  polarization ambiguity, and then the refined polarization information cues are used to guide 3DGS’s normal and spherical harmonic representation. This process achieves high-fidelity reflection separation and full-scene reconstruction without requiring environment maps or restrictive material assumptions. We demonstrate on public and self-collected datasets that PolarGuide-GSDR achieves state-of-the-art performance in specular reconstruction, normal estimation, and novel view synthesis, all while maintaining real-time rendering capabilities. To our knowledge, this is the first framework embedding polarization priors directly into 3DGS optimization, yielding superior interpretability and real-time performance for modeling complex reflective scenes.
\end{abstract}


\section{Introduction}
Multi-view high-quality reconstruction aims to recover the 3D geometry, texture, and even material properties of objects in a scene using images captured from multiple viewpoints. In recent years, the development of implicit surface representations~\cite{park2019deepsdf,sitzmann2020siren} and neural radiance fields (NeRF)~\cite{mildenhall2021nerf} has led to significant advances in geometric expressiveness and reconstruction quality. However, NeRF still suffers from limitations in training time and rendering efficiency. To address these issues, the 3D Gaussian Splatting (3DGS) method~\cite{kerbl2023gaussian} models the scene using sparsely distributed 3D Gaussian primitives instead of relying on a neural field, enabling high-quality rendering and real-time novel view synthesis while significantly accelerating training, thus alleviating NeRF’s performance bottlenecks. Despite their success in diffuse scenes, accurately modeling specular reflective surfaces remains a key challenge to achieving truly high-quality reconstruction.

To address the challenges of modeling specular reflections, neural rendering methods such as ReF-NeuS \cite{verbin2022refnerf}, NeRO \cite{liu2023nero}, and PANDORA \cite{dave2022pandora} incorporate inverse rendering with reflectance modeling to achieve photometric consistency on surface reflections. However, these approaches often suffer from long training times, low rendering efficiency, and limited generalizability. Although 3DGS leverages spherical harmonics (SH) to model view-dependent color variations, its low-order representation is insufficient to capture the high-frequency details characteristic of specular reflections. As a result, the model tends to fit specular highlights using fictitious Gaussian primitives during training, which can introduce geometric inaccuracies, particularly on non-planar surfaces \cite{kerbl2023gaussian}.

To improve the synthesis quality of specular reflection scenes, recent methods such as 3DGS with Deferred Reflection (3DGS-DR) have introduced deferred shading techniques \cite{ye2024deferredreflection}, while Ref-GS employs directional encoding \cite{zhang2025refgs} to enhance the expressive power of Gaussian primitives. However, due to the directional sensitivity of specular reflections and the entanglement with complex geometry, supervision relying solely on RGB images still struggles to achieve satisfactory results.

Polarization imaging, as a modality that simultaneously captures intensity, color, and polarization direction, provides physical priors closely related to surface normals and material reflectance properties. With the continuous advancement of polarization camera technology, polarization images can now be captured as conveniently as ordinary images. This has led to their widespread use in 3D reconstruction tasks involving reflective scenes. Current integration strategies mainly fall into two categories: one integrates polarization information into NeRF by solving for normals and material parameters through inverse rendering to assist reconstruction \cite{cao2023multi, han2024nersp, chen2024pisr, li2024neisf, li2025neisfplusplus, yang2024gnerp, gcchen2024pisr}; the other relies on polarization imaging models to separate specular and diffuse reflection components directly from images, serving as reconstruction priors \cite{cui2017polarimetric}. However, the former approach depends on object masks and entails high training overhead, limiting its applicability to large-scale or geometrically complex scenes; the latter suffers from azimuthal angle ambiguities, making the separated components difficult to use directly for supervision.

To this end, based on the physical model of polarization imaging \cite{cui2017polarimetric}, this work accurately separates specular and diffuse components from polarization images and introduces an ambiguity-correction strategy that utilizes surface normal priors from 3DGS to perform azimuthal angle correction on polarization normals, effectively resolving $\pi$ and $\pi/2$ ambiguities. Meanwhile, a deferred reflection mechanism is proposed, which jointly uses surface normals and specular images as supervision signals to constrain the geometric accuracy of polarization normals, significantly reducing interference from specular reflections. Building on this, we integrate polarization priors, 3DGS, and deferred reflection into a unified framework named Polarization-Guided Deferred Reflection with 3D Gaussian Splatting (PolarGuide-GSDR). Experiments demonstrate that our method maintains efficient reconstruction while significantly enhancing geometric and reflectance details in both specular and diffuse regions, validating its reconstruction capability and practical applicability in complex reflective scenes.	

Our contributions are summarized as follows.
\begin{itemize}
\item To address the strong assumption dependencies in polarization-NeRF inverse rendering, the reliance on environment maps in 3DGS deferred reflection, and the lack of reflection surface priors, we propose PolarGuide-GSDR, a forward-guided paradigm. This paradigm achieves reflection separation and full-scene reconstruction without requiring environment maps or strong assumptions through an image decoupling and categorical guidance mechanism.
\item To resolve the ambiguity in polarization-based normal estimation and mitigate specular reflection interference, we introduce an ambiguity-correction strategy based on cosine similarity and Degree of Linear Polarization (DoLP) thresholding. This approach leverages 3DGS geometric priors to dynamically correct polarization normal ambiguity, then employs the refined polarization information to guide 3DGS's reflection modeling and geometric optimization in reverse, establishing a bidirectional coupled iterative optimization process.
\item Confronting the limitations of existing polarization datasets sparse viewpoints and limited reflection information. we construct the first full-scene multi-view polarization dataset covering complex indoor and outdoor environments. Experiments demonstrate that PolarGuide-GSDR achieves leading performance on both public benchmarks and our self-collected dataset.
\end{itemize}

\section{Related Works}
\subsection{Specular Surface Reconstruction with NeRF}
In recent years, neural implicit representations \cite{xie2022neural} have driven the rapid advancement of neural inverse rendering techniques \cite{tewari2022advances}. These representations are applied in NeRF \cite{mildenhall2021nerf}, which implicitly models color and density to enable high-quality novel view synthesis. Since NeRF's introduction, numerous works have extended its volume rendering framework. For instance, methods like IDR \cite{yariv2020multiview}, NeuS \cite{wang2021neus}, and VolSDF \cite{yariv2021volume} integrate neural implicit representations to achieve high-precision 3D surface reconstruction from multi-view images. NeRF treats scene points' outgoing radiance as a mixture of material properties and illumination, research including NeuralPIL \cite{boss2021neural}, PhySG \cite{zhang2021physg}, and Ref-NeRF \cite{verbin2022ref}—focuses on disentangling reflectance and lighting components to improve physical consistency and interpretability.

Polarization, as a fundamental property of electromagnetic waves, provides rich geometric and material information through surface interactions and is widely used in inverse rendering tasks. PANDORA pioneered the integration of polarization into NeRF, employing coordinate-based MLPs to estimate surface normals along with diffuse/specular irradiance, followed by a simplified physical renderer to generate outgoing Stokes vectors \cite{dave2022pandora}. Subsequent developments include: NeRSP leveraging polarization angles to enhance geometric estimation and specular modeling under sparse views \cite{han2024nersp}; GNeRP incorporating polarization supervision and Gaussian normal modeling within SDFs, combined with a DoLP weighting strategy to suppress noise \cite{yang2024gnerp}; NeISF utilizing neural incident Stokes fields with differentiable polarization renderers to simulate multi-reflection propagation for improved geometry/material recovery \cite{li2024neisf}; NeISF++ extending these capabilities to conductive media scenarios \cite{li2025neisfplusplus}; and PISR enhances high-precision surface geometry reconstruction for textureless and reflective objects by incorporating a hash-grid accelerated neural SDF, though it does not extensively explore the rendering of reflective content \cite{gcchen2024pisr}. Despite progress, most methods remain NeRF-based, requiring auxiliary masks while suffering from low rendering efficiency, prolonged training, and limited applicability. Validation primarily occurs in idealized indoor environments, making real-world complex scenes challenging to handle. Therefore, achieving efficient and robust polarization-assisted rendering remains a critical open challenge.

\begin{figure*}[t]  
	\centering
	\includegraphics[width=0.98\linewidth]{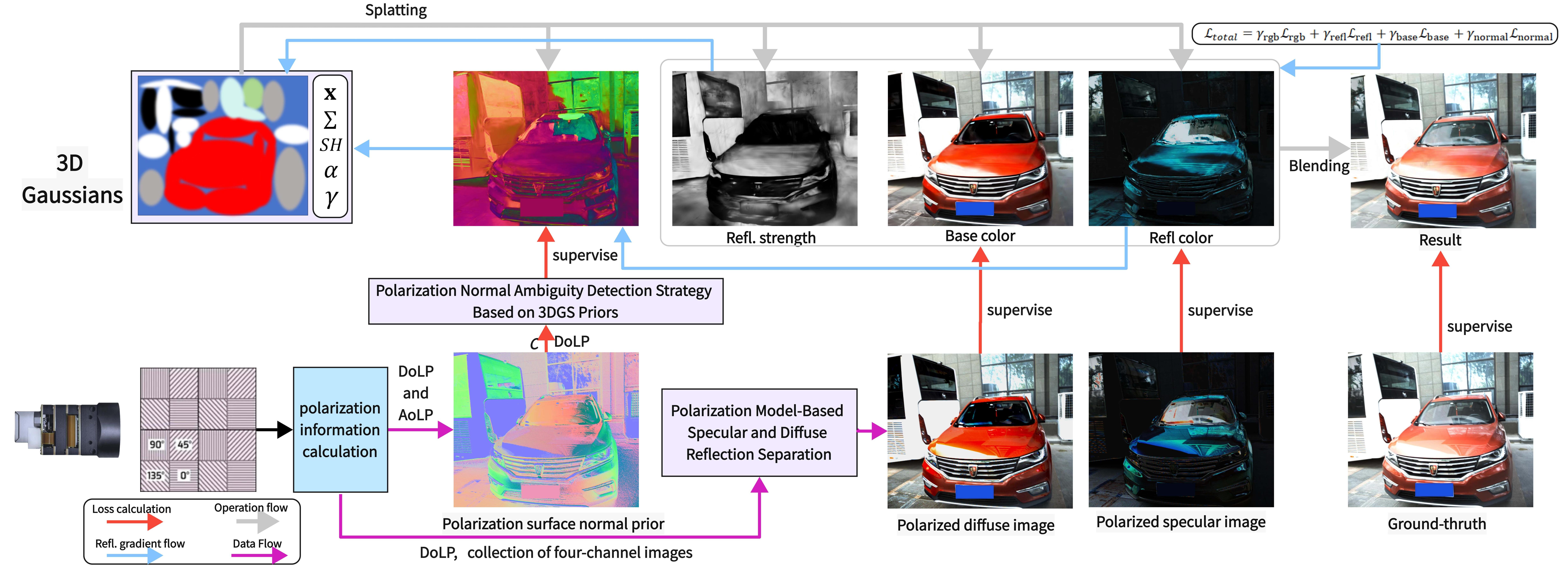}
	\caption{Overview of Polarguide-3DGS. First, polarization images are input to the system. The polarization information calculation module computes DoLP and AoLP, and extracts four-channel polarization images. Then, the Polarization Model-Based Specular and Diffuse Reflection Separation generates polarized specular and diffuse images as priors. And the initial polarization surface normal prior is estimated from DoLP and AoLP, then refined by the Polarization Normal Ambiguity-Correction Strategy Guided by 3DGS Priors to resolve normal estimation ambiguities. Finally, the polarized  Specular and Diffuse Reflection images, refined normals, and RGB ground truth jointly guide and supervise the pre-trained 3DGS model to enhance reflectance and geometry learning. During rendering, the estimated reflection strength map blends specular and diffuse outputs to produce realistic reflections.
	}
	\label{fig2}
\end{figure*}

\begin{figure*}[t]  
	\centering
	\includegraphics[width=\linewidth]{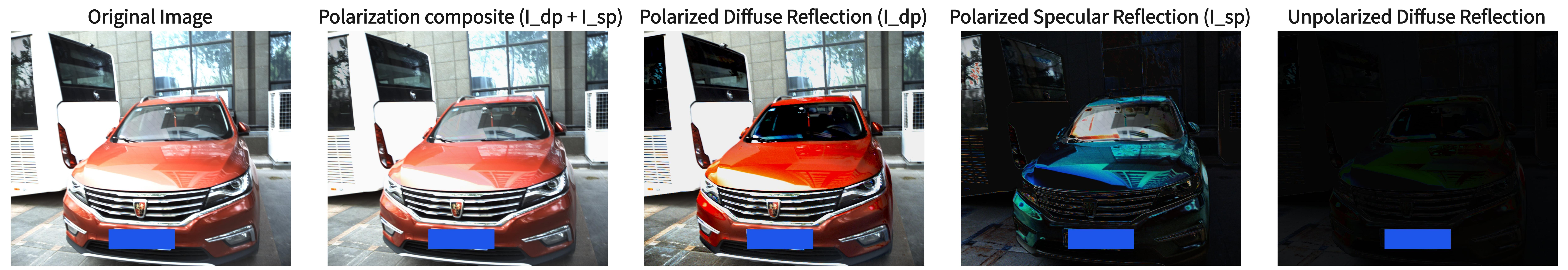}
	\caption{Separation results of specular and diffuse components based on polarized images.}
	\label{fig3}
\end{figure*}

\subsection{Specular Surface Reconstruction with Gaussian Splatting}
Recently, Kerbl et al. proposed the 3DGS method \cite{kerbl2023gaussian}, offering an attractive solution for novel view synthesis of 3D scenes. Compared to NeRF, 3DGS provides higher rendering efficiency, shorter training times, explicit geometric representation, and natural compatibility with traditional rendering strategies such as deferred reflection.

However, the presence of specular reflections remains a significant challenge for GS methods. Although 3DGS achieves view-dependent shading by employing SH for each Gaussian component, its limited directional frequency restricts its ability to accurately represent specular reflections, resulting in reflection distortions and geometric degradation. To address this issue, various recent extensions have been proposed to enhance specular reflection modeling capabilities. For instance, 3iGS introduces continuous tensor lighting fields and per-Gaussian BRDF factorization to decouple incident illumination and material response \cite{tang20243igs}; GaussianShader proposes a non-learned Gaussian directional encoding scheme to capture view dependence \cite{jiang2024gaussianshader}; 3DGS-DR improves reflection fitting by employing deferred reflection mechanisms and normal propagation training, enabling staged rendering \cite{ye2024deferredreflection}; furthermore, Ref-GS builds upon the 3DGS-DR framework by incorporating directional encoding and illumination factorization to better distinguish view and lighting direction information, enhancing reflection quality and geometric accuracy \cite{zhang2025refgs}.

Despite this, these methods lack explicit disentanglement of geometry, lighting, and materials, and it is difficult to accurately separate diffuse and specular reflections in complex reflective scenes. The splatting-based explicit rasterization in 3DGS limits its integration with physical priors such as polarization information. Polarization images, as passive optical signals, effectively provide geometric and material cues, particularly excelling in reflection separation and normal estimation. Based on this, we propose a Gaussian deferred reflection modeling method that integrates polarization priors, combining diffuse-specular separation, an ambiguity-correction strategy using cosine similarity and DoLP thresholds, and a deferred rendering strategy. This approach significantly improves geometry reconstruction and rendering stability in complex reflective scenes while maintaining 3DGS’s real-time performance.

\section{Method}
As shown in Figure~\ref{fig2}, the proposed PolarGuide-GSDR framework consists of three tightly integrated modules. The first, Polarization Model-Based Specular and Diffuse Reflection Separation, extracts specular and diffuse components and estimates initial surface normals, providing physical priors for reflection modeling. The second, Polarization Normal Ambiguity-correction strategy Based on 3DGS Priors, leverages DoLP thresholds, 3DGS geometry, and specular cues to guide and supervise the disambiguation of polarization normals. The third, Supervised Loss Design Integrating Polarization Information, incorporates the extracted priors to jointly guide and supervise the training of the deferred reflection and Gaussian rendering pipelines, enhancing geometric and appearance modeling in reflective scenes.

\subsection{Polarization Model-Based Specular and Diffuse Reflection Separation}
In 3D reflection modeling, distinguishing specular and diffuse components is crucial to improving geometric reconstruction and image synthesis quality. However, existing 3DGS-based approaches lack illumination modeling and reflection separation mechanisms, making it difficult to accurately reconstruct the spatial distribution of specular and diffuse reflections in real-world scenes. In contrast, polarization-assisted NeRF leverages view-dependent querying and inverse rendering to estimate material properties, but 3DGS lacks an explicit mechanism for view-dependent queries, limiting such strategies. To overcome this, we introduce a polarization-based physical model to separate specular and diffuse reflections at the image level using polarization images, providing stable and physically consistent supervisory priors for 3DGS and effectively enhancing geometry and appearance in specular regions.

The polarization characteristics of surface-reflected light are primarily determined by the polarized bidirectional reflectance distribution function (polarized BRDF) \cite{brayford2008physical, collin2020computation}. For most surfaces, the reflected light can typically be decomposed into three components: polarized specular reflection, polarized diffuse reflection, and unpolarized diffuse reflection \cite{cui2017polarimetric}. As illustrated in Figure~\ref{fig3}, the proportion of unpolarized diffuse reflection is negligible in most real-world scenarios and is therefore omitted in our modeling process.

We models the reflected light as two polarized components: the specular reflection component $S_{sp}$ and the diffuse reflection component $S_{dp}$, corresponding to Fresnel reflection and subsurface scattering, respectively. Both components can be observed using a linear polarizer placed in front of the camera. The polarization camera to capture polarized images. It simultaneously captures images at four different polarization angles:  $\left(0,\, \pi/4,\, \pi/2,\, 3\pi/4 \right)$, denoted as $I_{0^{\circ}}, \quad I_{45^{\circ}}, \quad I_{90^{\circ}}, \quad I_{135^{\circ}}$, respectively. Let the Stokes vector of the incident light be $S_{in} = \left[ S_0, S_1, S_2, S_3 \right]$, where 
\begin{eqnarray}
	S_0 &=& 0.5 \cdot \left(I_{0^\circ} + I_{45^\circ} + I_{90^\circ} + I_{135^\circ} \right) \nonumber \\
	S_1 &=& I_{0^\circ} - I_{90^\circ},\quad S_2 = I_{45^\circ} - I_{135^\circ}
\end{eqnarray}
 $S_3$ is unused. The Mueller matrix of a linear polarizer at angle $\theta$ is denoted as $M_{\mathrm{pol}}(\theta)$, while the Mueller matrices for Fresnel reflection and transmission are denoted as $M_R$ and $M_T$, respectively. The two polarization components can then be expressed as follows: $S_{s p}=M_{p o l}(\theta) M_{R} S_{i}, \quad S_{d p}=M_{p o l}(\theta) M_{T} S_{d}$. where \( S_d \) is the Stokes vector of the diffuse reflection. For efficient rendering-guided reflection separation, we assume \( S_d \approx S_i - S_{sp} \). This assumption provides an initial cue for reflection separation, and the approximation error will be compensated during the subsequent intensity map fusion stage.

By inverting the DoLP formula for polarized diffuse reflection in the Nayar model \cite{lei2022shape}, the zenith angle, which is the light incidence angle \(i\) on the object surface, can be computed. Combined with the surface refractive index $\mathrm{ref}_{id}$, the refraction angle $r$ is then obtained via Snell's law. The DoLP is calculated as follows:
\begin{eqnarray}
	DoLP=\frac{\sqrt{S_{1}^{2}+S_{2}^{2}}}{S_{0}}
\end{eqnarray}

Specular reflection image is $I_{sp}$:
\begin{eqnarray}
	I_{s p}\left(\phi_{p o l}\right)=\frac{I_{\max }^{s p}+I_{\min }^{s p}}{2}+\frac{I_{\max }^{s p}-I_{\min }^{s p}}{2} \cos (2 \theta) \label{eq3}
\end{eqnarray}
where $
I_{\max }^{s p}=\frac{S_{0}}{2}\left(\tan \alpha_{-}/\sin \alpha_{+}\right)^{2} \cos ^{2} \alpha_{-}, 
I_{\min }^{s p}=\frac{S_{0}}{2} \\ \left({\tan \alpha_{-}}/{\sin \alpha_{+}}\right)^{2} \cos ^{2} \alpha_{+}, 
\alpha_{ \pm}=i \pm r$.
$\theta$ denotes the angle between the polarization direction and the $x$-axis and $\phi_{\mathrm{pol}}$ denotes the channel angle of the polarizer.

Diffuse reflection image is $I_{d p}$:
\begin{equation}
	\begin{split}
		I_{d p}\left(\phi_{p o l}\right)=\frac{I_{\max }^{d p}+I_{\min }^{d p}}{2}+\frac{I_{\max }^{d p}-I_{\min }^{d p}}{2} \cos \left(2\left(\theta-\frac{\pi}{2}\right)\right)
		\label{eq4}
	\end{split}
\end{equation}
where 
$
I_{\max }^{d p}=\frac{S_{d}}{2} * \sin 2 i \sin 2 r /\left(\sin \alpha_{+} \cos \alpha_{-}\right)^{2},
I_{\min }^{d p}=\frac{S_{d}}{2} * \sin 2 i \sin 2 r * \cos ^{2} \alpha_{-}/\left(\sin \alpha_{+} \cos \alpha_{-}\right)^{2} 
$.

Based on the relationship between the polarization angle and the angle of incidence, the detailed derivation can be found in the supplementary materials and in \cite{cuipolarimetric}. However, the angle of incidence exhibits ambiguity during the reflection separation process, which in turn affects the accuracy of surface normal estimation. To address this, we propose a polarization normal ambiguity-correction strategy based on 3DGS priors, aiming to improve the stability and reliability of normal estimation.

\begin{table*}[ht!]  
	\centering
	\setlength{\tabcolsep}{4pt} 
	\renewcommand{\arraystretch}{1.2} 
	\begin{tabular}{ccccccc}
		\hline
		\multicolumn{1}{c}{Datasets} &  & GNeRP~\cite{yang2024gnerp} & 3DGS~\cite{kerbl2023gaussian} & 3DGS-DR~\cite{ye2024deferredreflection} & Ref-GS~\cite{zhang2025refgs} & PolarGuide-GSDR \\ \hline
		\multirow{4}{*}{\begin{tabular}[c]{@{}c@{}}Gnome\\ (Indoor)\end{tabular}} & PSNR~↑ & 17.65 & 19.37 & 21.13 & 21.65 & \textbf{22.54} \\
		& SSIM~↑ & 0.716 & 0.841 & 0.861 & 0.880 & \textbf{0.890} \\
		& LPIPS~↓ & 0.481 & 0.246 & 0.252 & 0.244 & \textbf{0.216} \\
		& FPS~↑ & 0.01 & 189.62 & 53.73 & 50.21 & 43.57 \\ \hline
		\multirow{4}{*}{\begin{tabular}[c]{@{}c@{}}Gundam\\ (Indoor)\end{tabular}} & PSNR~↑ & 15.42 & 22.78 & 22.93 & 22.85 & \textbf{23.32} \\
		& SSIM~↑ & 0.619 & 0.832 & \textbf{0.834} & 0.823 & 0.825 \\
		& LPIPS~↓ & 0.708 & 0.218 & \textbf{0.217} & 0.237 & 0.223 \\
		& FPS~↑ & 0.01 & 182.39 & 71.69 & 24.48 & 64.86 \\ \hline
		\multirow{4}{*}{\begin{tabular}[c]{@{}c@{}}Automotive and Glass\\ (Outdoor)\end{tabular}} & PSNR~↑ & 13.20 & 18.21 & 18.31 & 17.78 & \textbf{19.29} \\
		& SSIM~↑ & 0.595 & 0.762 & 0.763 & 0.754 & \textbf{0.774} \\
		& LPIPS~↓ & 0.595 & 0.340 & 0.343 & 0.362 & \textbf{0.339} \\
		& FPS(↑) & 0.01 & 279.78 & 118.30 & 43.23 & 104.63 \\ \hline
		\multirow{4}{*}{\begin{tabular}[c]{@{}c@{}}Black ceramic cup\\ (Indoor)\end{tabular}} & PSNR~↑ & 15.77 & 25.18 & 25.57 & 26.48 & \textbf{26.67} \\
		& SSIM~↑ & 0.667 & 0.889 & 0.886 & \textbf{0.902} & 0.894 \\
		& LPIPS~↓ & 0.386 & 0.199 & 0.199 & \textbf{0.186} & 0.191 \\
		& FPS~↑ & 0.01 & 255.50 & 108.83 & 36.70 & 81.52 \\ \hline
		\multirow{4}{*}{\begin{tabular}[c]{@{}c@{}}Stagnant water\\ (Outdoor)\end{tabular}} & PSNR~↑ & 17.55 & 22.65 & 23.00 & 23.01 & \textbf{23.51} \\
		& SSIM~↑ & 0.458 & 0.760 & 0.754 & 0.761 & \textbf{0.762} \\
		& LPIPS~↓ & 0.763 & \textbf{0.271} & 0.283 & 0.276 & 0.279 \\
		& FPS~↑ & 0.02 & 251.18 & 102.02 & 23.43 & 95.30 \\ \hline
	\end{tabular}
	\caption{Different Methods Compared on Reconstruction Quality and Rendering Efficiency under Real Indoor and Outdoor Scene Test Viewpoints}
	\label{tab1}
\end{table*}

\subsection{Polarization Normal Ambiguity-Correction Strategy Based on 3DGS Priors}
Based on the physical model of polarization, the angle of linear polarization(AoLP) corresponds to the azimuth angle $\sigma$, which is calculated as follows:
\begin{eqnarray}
	AoLP=\frac{1}{2} \operatorname{atan2}(S_2, S_1)
\end{eqnarray}
Meanwhile, the angle of incidence $i$ is calculated from the DoLP and the material's refractive index. By combining the azimuth angle $\sigma$ and the incidaence angle $i$, the polarized surface normal $\mathbf{n}_{\mathrm{pol}}$ can be derived.
\begin{eqnarray}
	n_{p o l}=\left[\begin{array}{l}
		n_{x} \\
		n_{y} \\
		n_{z}
	\end{array}\right]=\left[\begin{array}{c}
		\sin i \cdot \cos \sigma \\
		\sin i \cdot \sin \sigma \\
		\cos i
	\end{array}\right] \label{eq6}
\end{eqnarray}

However, there exists a periodic relationship between the AoLP and the normal azimuth in polarization data, inevitably inducing $\pi$ and $\pi/2$ directional ambiguities in polarization-derived normals. Additionally, the DoLP reflects the reliability of polarization information across different regions: in areas with high DoLP, the linear polarization characteristics of light are more pronounced, leading to more accurate estimation of the angle of polarization and zenith angle, making them suitable as references for surface normal supervision. In contrast, regions with low DoLP exhibit weaker polarization characteristics, resulting in greater uncertainty in normal estimation. It is worth noting that, since specular reflection is not excluded during the normal computation process, the polarized normals derived from Equation \ref{eq6} may still be affected by specular components.

Based on this, we introduce a threshold $\tau$ to supervise predicted normals only where DoLP is high, thereby improving physical consistency and training stability. Meanwhile, we construct a candidate set $\mathcal{C} = \{ \mathbf{n}_{pol}, -\mathbf{n}_{pol}, R_{\pi/2}(\mathbf{n}_{pol}), -R_{\pi/2}(\mathbf{n}_{pol}) \}$, where $R_{\pi/2}$ denotes a 90° rotation about the z-axis. As shown in Figure~\ref{fig2}, we combine the specular reflection image and surface normals to jointly guide and supervise the predicted normals of 3DGS, effectively removing the interference in polarization-derived normals caused by specular reflections.

\begin{figure*}[t]  
	\centering
	\includegraphics[width=\linewidth]{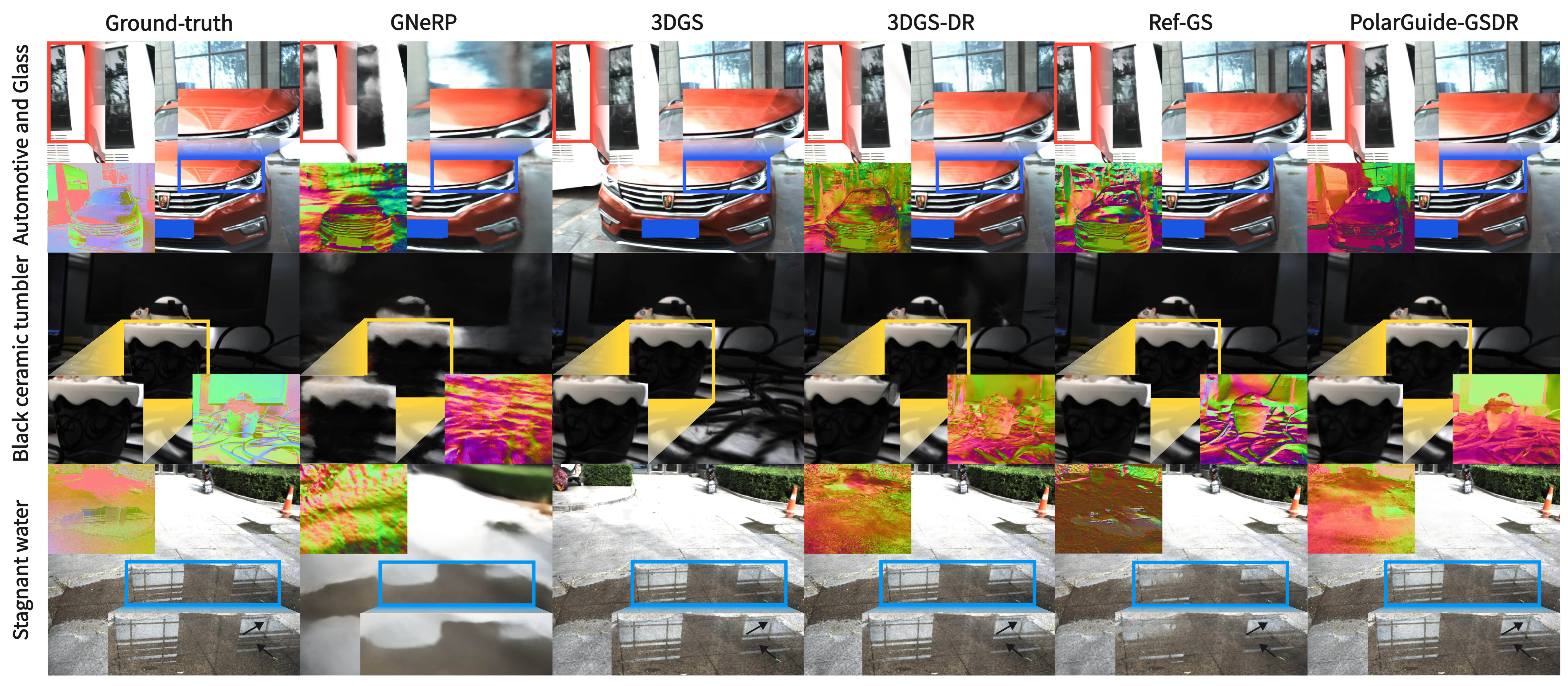}
	\caption{Comparison of Test Views Across Different Indoor and Outdoor Scenes (Note: Surface normals in GT are derived from polarization data, not measured by scanning, and are included to demonstrate the effectiveness of polarization-based normal extraction.)}
	\label{fig4}
\end{figure*}

\subsection{Supervised Loss Design Integrating Polarization Information}
In the 3DGS rendering pipeline, this work builds upon the 3DGS-DR method \cite{ye2024deferredreflection} to construct the rendering framework. This method employs a deferred rendering strategy that divides the rendering process into two streams: the Gaussian splatting pipeline, responsible for generating the basic spatial distribution and coarse colors; and the deferred reflection pipeline, which models specular reflection effects. These two streams are fused through a Gaussian scalar $r_i$ representing the specular reflection intensity to synthesize the final rendered image.

However, due to the lack of physical priors constraining specular reflection, diffuse reflection, and surface normals, this method is prone to generating spurious structures and unrealistic reflection behaviors during reflection field modeling. To address this issue, as shown in Figure~\ref{fig2}, we introduce a polarization-based supervision strategy. By leveraging the polarization priors computed from Equations~\ref{eq3}, \ref{eq4}, and \ref{eq6}, we provide physically consistent guidance for the specular component, diffuse component, and surface normals, thereby improving the physical plausibility and structural accuracy of the rendered results. To balance image quality and structural fidelity, the image loss is composed of the L1 loss and the structural similarity loss (D-SSIM).

In terms of loss function design, this work constructs the following four types of supervision terms:
\begin{itemize}
	\item Image reconstruction loss $\mathcal{L}_{rgb}$: \\
	\scalebox{0.93}{$
		\mathcal{L}_{\mathrm{rgb}} = (1 - \lambda_{\mathrm{rgb}}) \mathcal{L}_{1}(I_{\mathrm{final}}, I_{\mathrm{rgb}}) + \lambda_{\mathrm{rgb}} \mathcal{L}_{\text{D-SSIM}}(I_{\mathrm{final}}, I_{\mathrm{rgb}})
		$},\\
	$I_{\text{final}}$ represents the final image obtained by blending specular and diffuse reflections based on intensity, $I_{\mathrm{rgb}}$ denotes the ground-truth image.
	
	\item Specular reflection supervision loss $\mathcal{L}_{refl}$: \\
	\scalebox{0.93}{$\mathcal{L}_{\text {refl}}=(1-\lambda_{\mathrm{refl}}) \mathcal{L}_{1}\left(I_{\text {refl}}, I_{s p}\right)+\lambda_{\mathrm{refl}} \mathcal{L}_{\text{D-SSIM}}\left(I_{\text {refl}}, I_{sp}\right)$}
	, \\$I_{\text{refl}}$ denotes the rendered specular reflection image, $I_{\mathrm{sp}}$ is the polarized specular reflection image computed by Equation~\ref{eq3}.
	
	\item Diffuse reflection supervision loss $\mathcal{L}_{base}$: \\
	\scalebox{0.93}{$
		\mathcal{L}_{\text {base}}=\left(1-\lambda_{\text {base}}\right) \mathcal{L}_{1}\left(I_{\text {base}}, I_{dp}\right)+\lambda_{\text {base}} \mathcal{L}_{\text{D-SSIM}}\left(I_{\text {base}}, I_{dp}\right)
		$}, \\$I_{\text{base}}$ denotes the rendered diffuse reflection image, $I_{\mathrm{dp}}$ is the polarized diffuse reflection image computed by Equation~\ref{eq4}.
	\item Surface normal supervision loss $\mathcal{L}_{normal}$: \\
	\scalebox{0.93}{$
		\mathcal{L}_{\text {normal}}=\frac{1}{N} \sum_{x, y, z} 1_{\text {DoLP}>\tau} \cdot \min _{c \in \mathcal{C}}\left[1-\cos \left(n_{\text {pred }}, c \right)\right]
		$}\\, when $1_{\text {DoLP}>\tau}=\left\{\begin{array}{ll}
		1, & d>\tau \\
		0, & d \leq \tau
	\end{array}\right.$\\
	$\mathbf{n}_{\mathrm{pred}}$ denotes the surface normal predicted by 3DGS. $\mathcal{C}$ is the candidate set of polarization-based normals.
	
\end{itemize}
The overall multi-term joint loss function is expressed as:
\begin{equation}
	\begin{split}
		\mathcal{L}_{\textrm{total}} =
		\eta_{\textrm{rgb}} \mathcal{L}_{\textrm{rgb}} +
		\eta_{\textrm{refl}} \mathcal{L}_{\textrm{refl}} +
		\eta_{\textrm{base}} \mathcal{L}_{\textrm{base}} + \eta_{\textrm{normal}} \mathcal{L}_{\textrm{normal}}
	\end{split}
\end{equation}

Polarization cues are used exclusively for supervision during training and are not required at inference time. The weights of each loss term ($\lambda$ and $\eta$) are determined through experimental tuning to achieve an optimal balance between image reconstruction quality, the accuracy of specular and diffuse reflection modeling, and surface geometric consistency. (Comprehensive training procedures and parameter configurations are provided in the appendix and supplementary materials.)

\section{Experiments and Analysis}
Here, we conducted quantitative and qualitative comparisons of PolarGuide-GSDR with other state-of-the-art (SOTA) methods, including 3DGS-DR and Ref-GS, on a series of real-world datasets. We also performed an ablation study to verify the effectiveness of the proposed method.

\textbf{Dataset}. To validate the performance in real-world scenarios, we first tested on public datasets: the Gnome dataset from PANDORA \cite{dave2022pandora} and the Gundam (PISR-Figure dataset) dataset from PISR \cite{gcchen2024pisr}. However, their limited viewpoint coverage and weak reflective characteristics constrain comprehensive evaluation. We therefore constructed a more representative polarization dataset covering diverse indoor/outdoor scenes with varied reflective properties. Our new dataset includes: 1) Outdoor automotive and glass scene (strong specular reflections); 2) Indoor black ceramic cup scene; 3) Outdoor stagnant water scene (single-material surface, limited specular reflections).

\textbf{Baseline}. We compare with GNeRP~\cite{yang2024gnerp} (polarization and NeRF-based) and Gaussian-based methods 3DGS~\cite{kerbl2023gaussian}, 3DGS-DR~\cite{ye2024deferredreflection}, and Ref-GS~\cite{zhang2025refgs} using the same data and training setup.

\textbf{Evaluation Metrics}. We employ three widely adopted metrics: Peak Signal-to-Noise Ratio (PSNR) for image fidelity, Structural Similarity Index (SSIM) for structural consistency, and Learned Perceptual Image Patch Similarity (LPIPS) which aligns with human visual perception.

\textbf{Experimental Details}. For details, please refer to the supplementary material.

\subsection{Quantitative Experiments}
As shown in Table~\ref{tab1}, quantitative comparisons across five real-world datasets demonstrate that our method achieves significant advantages in image quality. By integrating polarization cues into the 3DGS-based deferred reflection framework, our approach enables high-quality geometry and reflection reconstruction across diverse complex scenes. Specifically, PSNR improves by approximately 1 dB on the public Gnome dataset and the self-captured Automotive and Glass and Black Ceramic Cup datasets, while gains of about 0.5 dB are observed on the Gundam and Stagnant Water datasets.

Further analysis reveals that the notable PSNR gains on Automotive and Glass and Black Ceramic Cup stem from their rich reflective content. In contrast, the Gnome dataset suffers from sparse views, poor lighting, and limited reflections, producing severe artifacts in all GS-based methods after COLMAP \cite{schonberger2016structure} initialization. Its modest PSNR improvement largely stems from minor artifact reduction in some images. Similarly, Gundam's sparse views and weak reflections restrict PSNR gains. Though Stagnant Water has limited reflective areas as an outdoor scene with modest gains, our method still enhances its reconstruction accuracy. As our method targets reflective regions, overall PSNR improvement is not the only metric; recovering reflective details remains more critical.

\begin{figure}[!htbp]
	\centering
	\includegraphics[width=\linewidth]{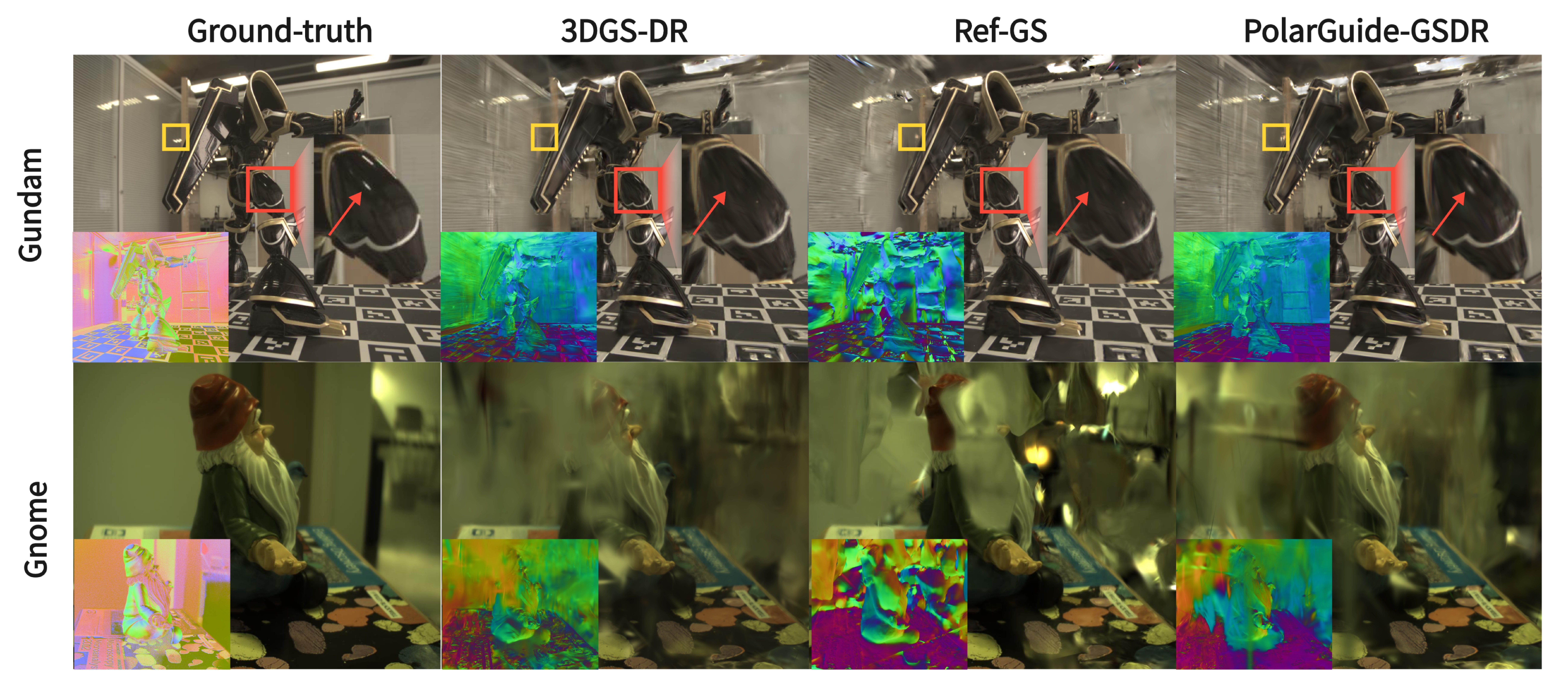}
	\caption{Comparison of Test Views on Public Datasets}
	\label{fig8}
\end{figure}

\subsection{Qualitative Experiments}
The proposed PolarGuide-GSDR maintains high-quality reconstruction while achieving rendering speed comparable to 3DGS-DR, significantly outperforming Ref-GS and GNeRP, while maintaining real-time rendering capabilities. As shown in Figure~\ref{fig4}, our method produces clear and accurate details in reflective regions: in the Automotive and Glass scene, the tree shadows on the left car window and the building reflections integrated with geometry on the red car hood are faithfully reconstructed; similarly, the reflection details in the marked areas of the Stagnant Water scene are accurately recovered. As shown in Figure~\ref{fig8}, the proposed method achieves favorable results on public datasets, with Gundam demonstrating clearer reconstruction of lighting reflections and Gnome showing moderately mitigated artifacts. However, due to space constraints, more detailed rendering comparisons, training procedures, and reproduction weights will be provided in the supplementary material.

To further evaluate reconstruction quality, we present surface normal estimation results of various methods in Figure~\ref{fig4}. Significant differences exist: GNeRP performs poorly and is not elaborated here; 3DGS-DR lacks explicit normal supervision, causing pronounced noise and disorganized surface structures; Ref-GS performs reasonably well but shows insufficient smoothness on the Automotive and Glass dataset, distortion in the Black Ceramic Tumbler’s background monitor region, and noticeable normal noise in the Stagnant Water dataset. In contrast, PolarGuide-GSDR introduces polarization-derived surface normals as priors and combines specular reflection supervision with a DoLP mask, providing effective constraints on normal estimation and suppressing specular interference. This results in smoother, more accurate surface normals and improved reconstruction quality in specular regions.

\begin{figure}[!htbp]
	\centering
	\includegraphics[width=\linewidth]{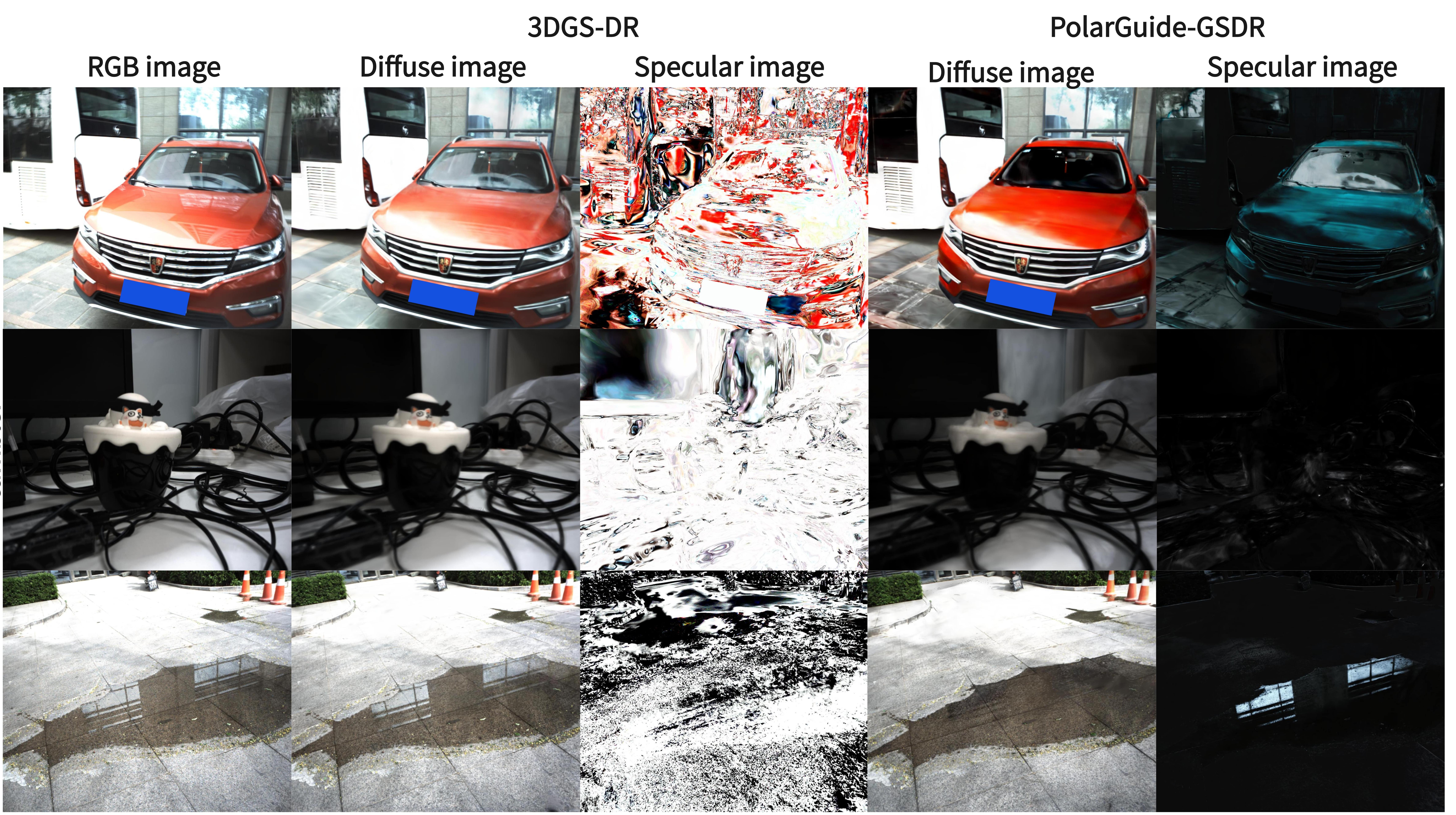}
	\caption{Comparison of Specular and Diffuse Reflection Space Reconstruction Results}
	\label{fig5}
\end{figure}

Simultaneously, further improvement in reconstruction accuracy depends on precise modeling of specular and diffuse reflection spaces. As shown in Figure~\ref{fig5}, the traditional 3DGS-DR method struggles to effectively disentangle reflection spaces in real-world scenes due to the lack of prior information on reflection components. Consequently, its Base Color output closely resembles the final rendered image, and the specular reflection component obtained via deferred reflection contributes only marginally. In contrast, the proposed PolarGuide-GSDR method leverages polarization information to achieve prior-based separation of reflection spaces: it first utilizes the physical properties of polarization to preliminarily decompose specular and diffuse reflection images, providing effective constraints for reconstruction; subsequently, the separated components are combined with optimized surface normals to collaboratively guide and supervise the modeling of both surface normals and reflection spaces. Experimental results demonstrate that this method achieves high-precision reconstruction of reflection components across most real-world scenes, significantly enhancing the physical realism of material representation.

\begin{table}[]
	\small
	\setlength{\tabcolsep}{1pt} 
	\renewcommand{\arraystretch}{1.2} 
	\begin{tabular}{cccccc}
		\hline
		\multicolumn{1}{c}{Datasets} &                                                                           & \begin{tabular}[c]{@{}c@{}}3DGS-DR\\(Baseline)\end{tabular} & \begin{tabular}[c]{@{}c@{}}Ours only \\ Polar-\\Normal\end{tabular} & \begin{tabular}[c]{@{}c@{}}Ours w/o \\ Polar-\\Normal\end{tabular} & \begin{tabular}[c]{@{}c@{}}PolarGuide\\ -GSDR\end{tabular} \\ \hline
		\multirow{4}{*}{\begin{tabular}[c]{@{}c@{}}Gnome\\ (Indoor)\end{tabular}}   & PSNR~↑   & 21.13                                              & 19.687                                                                                         & 19.26                                                                                   & \textbf{22.54}                                                      \\
		& SSIM~↑   & 0.861                                              & 0.832                                                                                         & 0.820                                                                                  & \textbf{0.890}                                                      \\
		& LPIPS~↓ & 0.252                                              & 0.282                                                                                         & 0.275                                                                                  & \textbf{0.216}                                                      \\
		& FPS~↑    & 53.73                                             & 54.55                                                                                        & 53.68                                                                                 & 43.57                                                     \\ 
		\hline
		\multirow{4}{*}{\begin{tabular}[c]{@{}c@{}}Gundam\\ (Indoor)\end{tabular}}   & PSNR~↑   & 22.93                                              & 22.97                                                                                         & 22.93                                                                                   & \textbf{23.32}                                                      \\
		& SSIM~↑   & \textbf{0.834}                                              & 0.826                                                                                         & 0.825                                                                                   & 0.825                                                      \\
		& LPIPS~↓ & \textbf{0.218}                                              & 0.224                                                                                         & 0.227                                                                                   & 0.223                                                      \\
		& FPS~↑    & 71.69                                             & 66.03                                                                                        & 78.72                                                                                  & 64.86                                                     \\ 
		\hline
		\multirow{4}{*}{\begin{tabular}[c]{@{}c@{}}Automotive\\ and Glass\\ (Outdoor)\end{tabular}}   & PSNR~↑   & 18.31                                              & 17.84                                                                                         & 18.74                                                                                   & \textbf{19.27}                                                      \\
		& SSIM~↑   & 0.763                                              & 0.748                                                                                         & 0.764                                                                                   & \textbf{0.774}                                                      \\
		& LPIPS~↓ & 0.343                                              & 0.358                                                                                         & 0.345                                                                                   & \textbf{0.339}                                                      \\
		& FPS~↑    & 118.30                                             & 115.89                                                                                        & 116.68                                                                                  & 104.63                                                     \\ \hline
		\multirow{4}{*}{\begin{tabular}[c]{@{}c@{}}Black\\ ceramic\\ cup\\ (indoor)\end{tabular}} & PSNR~↑   & 25.57                                              & 24.98                                                                                         & 25.20                                                                                   & \textbf{26.67}                                                      \\
		& SSIM~↑   & 0.886                                              & 0.883                                                                                         & 0.882                                                                                   & \textbf{0.895}                                                      \\
		& LPIPS~↓ & 0.199                                              & 0.201                                                                                         & 0.203                                                                                   & \textbf{0.192}                                                      \\
		& FPS~↑    & 108.83                                             & 91.66                                                                                         & 84.02                                                                                   & 81.52                                                      \\ \hline
		\multirow{4}{*}{\begin{tabular}[c]{@{}c@{}}Stagnant\\ water\\ (Outdoor)\end{tabular}}         & PSNR~↑   & 23.00                                              & 22.61                                                                                         & 22.67                                                                                   & \textbf{23.32}                                                      \\
		& SSIM~↑   & 0.754                                              & 0.755                                                                                         & 0.753                                                                                   & \textbf{0.756}                                                      \\
		& LPIPS~↓ & \textbf{0.283}                                              & 0.286                                                                                         & 0.292                                                                                   & 0.289                                                      \\
		& FPS~↑    & 102.02                                              & 101.41                                                                                        & 84.99                                                                                   & 101.59                                                     \\ \hline
	\end{tabular}
	\caption{Ablation Study of PolarGuide-GSDR.}
	\label{tab2}
\end{table}

\subsection{Ablation Study}
Since specular reflection reconstruction in the deferred reflection module depends on the joint guidance and supervision of both the surface normal map and the specular reflection image, relying on only one of them results in incomplete supervision, leading to ambiguities and reduced reconstruction accuracy. In certain scenarios, the performance even lags behind the 3DGS-DR baseline without polarization information (as shown in Table \ref{tab2}). On one hand, supervision from specular images effectively suppresses highlight-induced errors in polarization normals. On the other hand, normal supervision offers geometric priors that help accurately localize and reconstruct highlight structures. These two supervision signals exhibit strong complementarity. Ablation studies show that when only one supervision branch is retained during training, overall performance degrades, and in some cases, even falls below that of the 3DGS-DR baseline, confirming the necessity of joint supervision from both polarization normals and specular reflection images for high-quality reflective reconstruction.

\section{Conclusion}
We propose PolarGuide-GSDR, a 3D Gaussian Splatting Driven by Polarization Priors and Deferred Reflection for Real-World Reflective Scenes method. It significantly improves the clarity and accuracy of specular reflections while maintaining comparable real-time frame rates. By combining polarization cues with a DoLP-based masking mechanism, it effectively resolves ambiguities in normal estimation and achieves precise reconstruction of both specular and diffuse components. Although assuming a fixed refractive index of 1.5 for non-conductive materials, PolarGuide-GSDR demonstrates excellent generalization capability, enabling high-precision full-scene reconstruction even for conductive materials like automotive surfaces. This successfully overcomes the strong viewpoint/illumination dependencies and single-object modeling limitations of traditional polarization-based NeRF methods.

\bibliographystyle{IEEEtran}
\bibliography{main}


\section{Supplementary Material}
To support the theoretical foundation of our proposed method, we include the detailed derivation of the polarization-based specular and diffuse reflection separation model in this section.
\subsection{Polarization Model-Based Specular and Diffuse Reflection Separation}

According to~\cite{cui2017polarimetric}, the polarization characteristics of reflected light are primarily determined by the surface material’s polarized bidirectional reflectance distribution function (polarized BRDF)~\cite{brayford2008physical, collin2020computation}. For most surfaces, the reflected light can typically be decomposed into three components: polarized specular reflection, polarized diffuse reflection, and unpolarized diffuse reflection. 

We models the reflected light as two polarized components: the specular reflection component $S_{sp}$ and the diffuse reflection component $S_{dp}$, corresponding to Fresnel reflection and subsurface scattering, respectively. Both components can be observed using a linear polarizer placed in front of the camera. The polarization camera to capture polarized images. It simultaneously captures images at four different polarization angles:  $\left(0,\, \pi/4,\, \pi/2,\, 3\pi/4 \right)$, denoted as $I_{0^{\circ}}, \quad I_{45^{\circ}}, \quad I_{90^{\circ}}, \quad I_{135^{\circ}}$, respectively. Let the Stokes vector of the incident light be $S_{in} = \left[ S_0, S_1, S_2 \right]$
, where 
\begin{eqnarray}
	S_0 &=& 0.5 \cdot \left(I_{0^\circ} + I_{45^\circ} + I_{90^\circ} + I_{135^\circ} \right) \nonumber \\
	S_1 &=& I_{0^\circ} - I_{90^\circ} \nonumber \\
	S_2 &=& I_{45^\circ} - I_{135^\circ}
\end{eqnarray}

The Mueller matrix of a linear polarizer at angle $\theta$ is denoted as $M_{\mathrm{pol}}(\theta)$, while the Mueller matrices for Fresnel reflection and transmission are denoted as $M_R$ and $M_T$, respectively. The two polarization components can then be expressed as follows:
\begin{eqnarray}
	S_{s p}=M_{p o l}(\theta) M_{R} S_{i} \nonumber\\
	S_{d p}=M_{p o l}(\theta) M_{T} S_{d}
\end{eqnarray}	
Where \( S_d \) is the Stokes vector of the diffuse reflection. To achieve efficient reflection separation for rendering guidance, we assume \( S_d \approx S_i - S_{sp} \). This assumption provides an initial cue for reflection separation, and the approximation error will be compensated during the subsequent intensity map fusion stage.

By inverting the DoLP formula for polarized diffuse reflection in the Nayar model \cite{lei2022shape}, the zenith angle, which is the light incidence angle \(i\) on the object surface, can be computed. Combined with the surface refractive index $\mathrm{ref}_{id}$, the refraction angle $r$ is then obtained via Snell's law. The DoLP is calculated as follows:
\begin{eqnarray}
	DoLP=\frac{\sqrt{S_{1}^{2}+S_{2}^{2}}}{S_{0}}
\end{eqnarray}

Let \( d = \mathrm{DoLP} \), \( k = \mathrm{ref}_{\mathrm{id}} \), and $
A = 2(1 - d) - (1 + d)\left(k^2 + \frac{1}{k^2}\right), \quad
B = 4d, \quad
C = 1 + k^2, \quad $
$
D = 1 - k^2.
$
Then, the incident angle $i$ can be derived as:
{\small
	\begin{equation}
		i = \arcsin \left( \sqrt{ \frac{ -B C (A + B) - \sqrt{ C^{2}(A + B)^{2} - D^{2}(A - B)^{2} } }{ 2(A^{2} - B^{2}) } } \right) \label{eq3_1}
	\end{equation}
}

Subsequently, based on Snell's law, the refraction angle \( r \) is computed using the refractive index and the incident angle \( i \) derived in Equation~\ref{eq3_1}:

Specular reflection image is $I_{sp}$:
\begin{eqnarray}
	I_{s p}\left(\phi_{p o l}\right)=\frac{I_{\max }^{s p}+I_{\min }^{s p}}{2}+\frac{I_{\max }^{s p}-I_{\min }^{s p}}{2} \cos (2 \theta) 
\end{eqnarray}
where 
\begin{eqnarray}
	I_{\max }^{s p}&=&\frac{S_{0}}{2}\left(\frac{\tan \alpha_{-}}{\sin \alpha_{+}}\right)^{2} \cos ^{2} \alpha_{-} \nonumber\\
	I_{\min }^{s p}&=&\frac{S_{0}}{2}\left(\frac{\tan \alpha_{-}}{\sin \alpha_{+}}\right)^{2} \cos ^{2} \alpha_{+} \nonumber \\
	\alpha_{ \pm}&=&i \pm 	r
\end{eqnarray}
$\theta$ denotes the angle between the polarization direction and the $x$-axis and $\phi_{\mathrm{pol}}$ denotes the channel angle of the polarizer.

Diffuse reflection image is $I_{d p}$:
\begin{eqnarray}
	I_{d p}\left(\phi_{p o l}\right)=\frac{I_{\max }^{d p}+I_{\min }^{d p}}{2}+\frac{I_{\max }^{d p}-I_{\min }^{d p}}{2} \cos (2(\theta-\pi / 2))
\end{eqnarray}
where 
\begin{eqnarray}
	I_{\max }^{dp}&=&\frac{S_{d}}{2} \frac{\sin 2 i \sin 2 r}{\left(\sin \alpha_{+} \cos \alpha_{-}\right)^{2}} \nonumber\\
	I_{\min }^{dp}&=&\frac{S_{d}}{2} \frac{\sin 2 i \sin 2 r}{\left(\sin \alpha_{+} \cos \alpha_{-}\right)^{2}} \cos ^{2} \alpha_{-}
\end{eqnarray}
A more detailed derivation process can be found in~\cite{cuipolarimetric}.

\subsection{Implementation Details of PolarGuide-GSDR}

\textbf{Experimental Details}.  We use COLMAP \cite{schonberger2016structure} for Gaussian initialization. All experiments in this study were conducted on the AutoDL cloud computing platform. Except for the GNeRP method, all other methods used a standard hardware configuration consisting of 30 GB of system RAM, a 50 GB data disk, a single NVIDIA GeForce RTX 4090 (24 GB) GPU, and a CPU—either an Intel Xeon Platinum 8352V with 16 vCPUs or an Intel Xeon Platinum 8481C with 25 vCPUs. Minor variations in CPU models across instances arose from parallel experiments on multiple datasets; however, all experiments for a given dataset were performed on the same instance to ensure consistency. Due to its higher GPU memory requirements, the GNeRP method was trained on two RTX 4090 (24 GB) GPUs, with a similar CPU configuration. The Python environment for all methods remained consistent with that used in the 3DGS baseline. 

\textbf{Training Process} As described in the methodology section of the main text, surface normals estimated from polarization cues exhibit ambiguity. To mitigate this uncertainty, we first perform initial training of the 3DGS and deferred reflectance model for approximately 10,000 iterations. Based on this initialization, the polarized surface normals are disambiguated. Subsequently, the corrected polarized surface normals, specular reflection images, and diffuse reflection images are used to guide and supervise the model training. Specifically, the diffuse reflection images supervise the diffuse component in both 3DGS and the deferred shading model, while the specular reflection images and surface normals primarily supervise the reconstruction of the specular reflection space in the deferred shading model. Constrained by the inherent difficulty in fully eliminating polarization ambiguity and separation inaccuracies near vertical reflection angles, we adopt a phased training strategy to guide the model toward gradual convergence. The detailed procedure is as follows:
\begin{itemize}
	\item Initial reconstruction using 3DGS for approximately 10,000 iterations.
	\item Introduction of corrected polarization priors to jointly supervise 3DGS and the deferred reflectance model for about 30,000–80,000 iterations.
	\item Final optimization using RGB images and Polarization normal for end-to-end refinement of the rendered output, lasting roughly 10,000 iterations.
\end{itemize}
\begin{figure*}[!ht]  
	\centering
	\includegraphics[width=\linewidth]{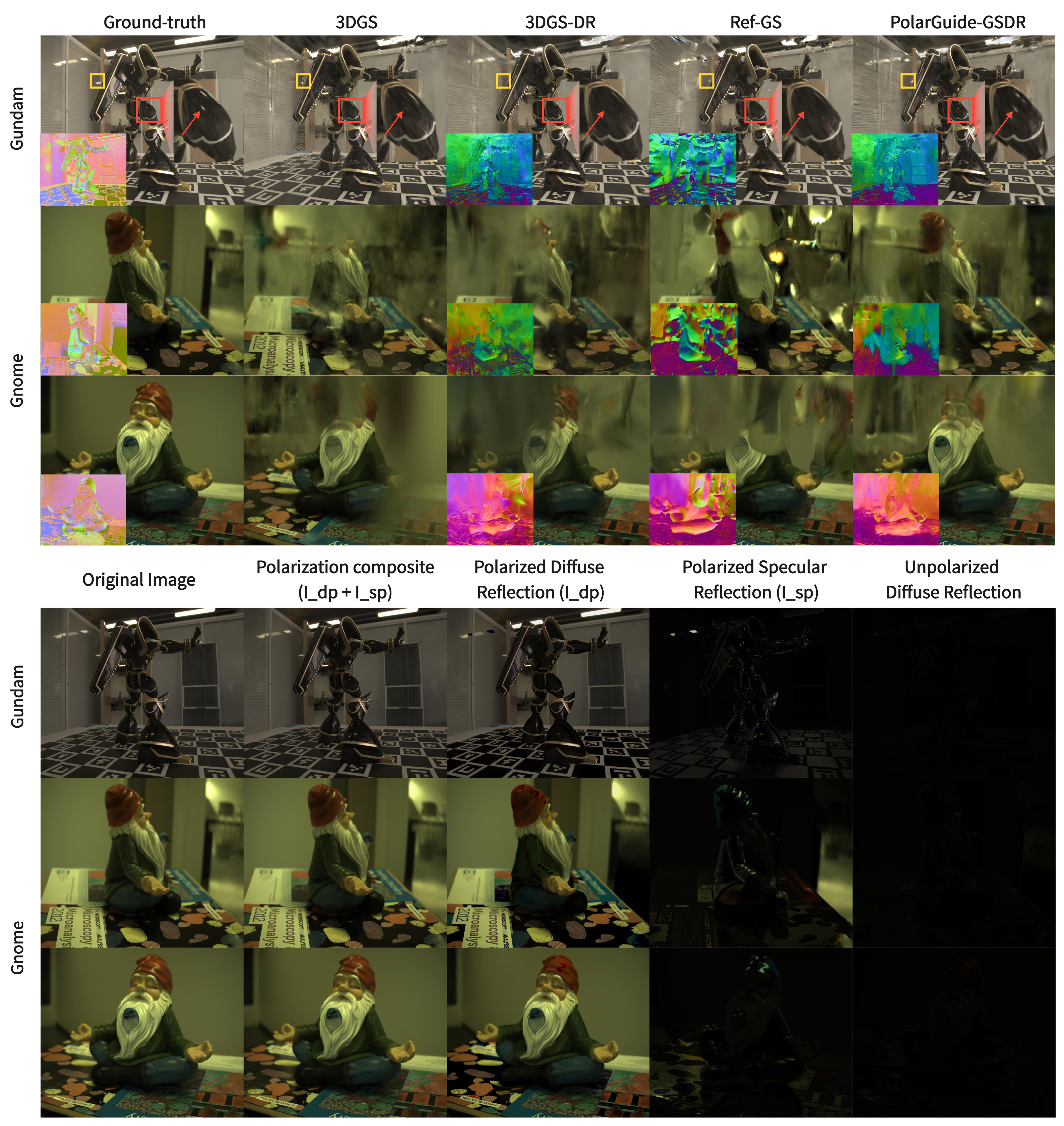}
	\caption{Comparison Test Views of Reconstruction Results and Reflection Separation on the Gnome~\cite{dave2022pandora} and Gundam~\cite{chen2024pisr} Datasets)}
	\label{fig7}
\end{figure*}
The total number of training iterations is set with reference to the 97,000 iterations used by 3DGS-DR in real-world scenes. For datasets with sparse viewpoints, we moderately reduce the total iteration count to prevent overfitting, while maintaining consistency with the iteration settings of 3DGS-DR. The hyperparameters $\lambda$ and $\eta$ mentioned in the main text are adjusted slightly across datasets; their specific values are provided in the configuration files included in the supplementary material.

Thus our method demonstrates significant advantages over prior works, such as NeISF++ \cite{li2025neisfplusplus}, particularly in its handling of material properties and its applicability to complex, real-world scenes.

The key distinction lies in our treatment of the refractive index. While methods like NeISF++ require the precise estimation of spatially-varying refractive indices, often involving additional neural fields like SDFs and typically limiting application to single-object, uniform-material scenes, we employ a fixed refractive index (e.g., 1.5) as part of a robust, phased training strategy.

This strategy renders our method largely agnostic to the specific value of the refractive index. The fixed refractive index is used only for an initial, coarse separation of specular and diffuse reflections. These coarsely separated components then serve as stable guidance to bootstrap the 3D reconstruction process in the early stages. Crucially, the final reconstruction quality is refined through end-to-end optimization supervised primarily by RGB images and polarization normals. This phased approach effectively decouples the final output from the inaccuracies of the initial coarse separation.

Consequently, our framework bypasses the challenging and often ill-posed problem of per-pixel refractive index estimation. This not only simplifies the system architecture by eliminating the need for complex networks to represent refractive indices but also enables the successful reconstruction of full, complex scenes containing objects with diverse materials—a scenario where methods requiring precise refractive index estimation typically fail.

\begin{figure*}[!ht]  
	\centering
	\includegraphics[width=\linewidth]{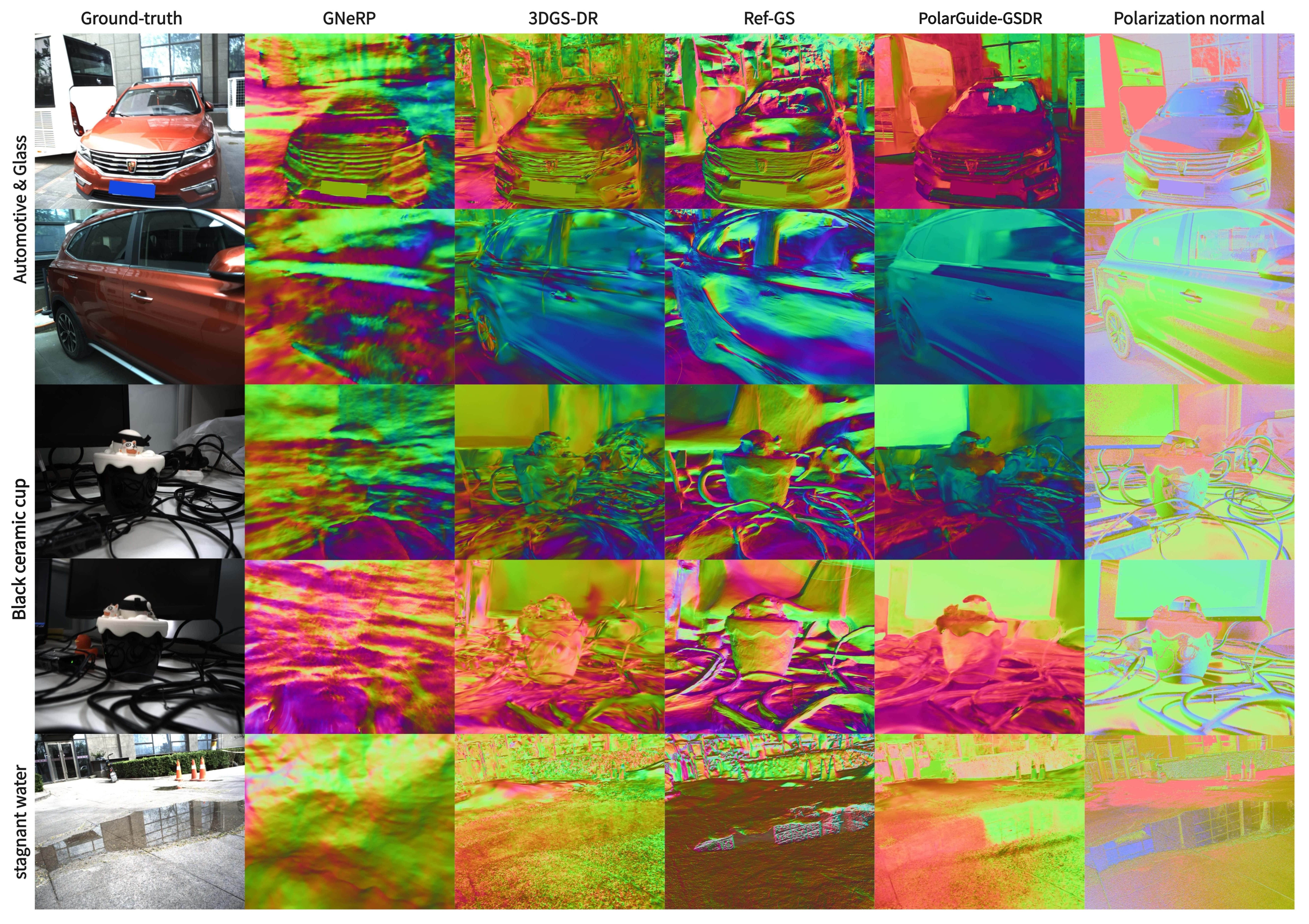}
	\caption{Surface Normals Comparison Chart of the Test Set in the Self-Collected Dataset}
	\label{fig9}
\end{figure*}

\textbf{Self-Collected Dataset Construction} For the self-collected dataset used in this study, image acquisition was performed with a Sony color polarization camera. In the absence of such specialized equipment, researchers can alternatively capture data by attaching a polarizing filter to a standard color RGB camera. Raw data for polarization information calculation can be obtained by rotating the filter and capturing images at four specific angles: 0°, 45°, 90°, and 135°.

\textbf{Training Time Cost} Among all compared algorithms, 3DGS requires the shortest training time, completing in under 30 minutes on most datasets. Due to an increased number of training iterations, 3DGS-DR takes approximately twice as long as 3DGS, typically around one hour. Although Ref‑GS is set to 30,000 iterations, its training time is slightly longer than that of 3DGS‑DR. Our proposed PolarGuide‑GSDR method employs polarization cues to jointly guide and supervise the 3DGS and deferred reflection models. The incorporation of additional polarization-based supervision introduces computational overhead, leading to a training time approximately 30\% longer than that of 3DGS‑DR, yet it remains comparable to the time required by Ref‑GS. Given that GNeRP demands significantly longer training—approximately 10 times that of 3DGS—it is excluded from detailed comparison here.

\subsection{Supplementary Comparison of Results on Public Datasets}
As shown in Figure \ref{fig7}, our method achieves a significant improvement in reconstruction quality on the Gundam dataset, particularly in restoring complex lighting and material reflections. This improvement is largely attributed to the effective separation of specular and diffuse reflections achieved through polarization imaging. This prior guides the model to separately reconstruct and fuse the two reflection components, leading to the synthesis of more photorealistic novel views. (Note: Since GNeRP produces severely blurred results when input masks are unavailable, we have excluded its corresponding quantitative comparisons to maintain valid analysis.)

However, the sparse nature of both datasets inevitably introduces some artifacts in the reconstructions. For instance, the Gnome dataset contains only 35 images in total (with 5 reserved for testing). To mitigate these artifacts, we separately reconstruct the diffuse and specular reflection spaces using polarization priors and introduce polarization-derived surface normal constraints. This approach significantly reduces artifacts and enhances the robustness of the reconstruction.

Due to space limitations in the main text and to facilitate intuitive correlation between the reconstructed surface normals and the original images, we have integrated them into the same figure. To more clearly demonstrate the comparative effectiveness of the normal reconstruction, Figure \ref{fig9} provides supplementary multi-view detailed illustrations. As shown, our proposed PolarGuide-GSDR method effectively suppresses noise in surface normal reconstruction while maintaining excellent detail characteristics: in complex curved areas, it avoids over-smoothing or over-sharpening of normals, restoring authentic geometric continuity; in weak-texture regions (such as stagnant water surfaces), it still reconstructs reasonable normal variations. PolarGuide-GSDR demonstrates significant performance improvements in both visual quality and geometric accuracy, fully validating the effectiveness of the polarization normal prior information.
\end{document}